\documentclass[review]{elsarticle}
\usepackage{amsmath}
\usepackage{algorithmic}
\usepackage{array}
\usepackage{lineno,hyperref}
\usepackage{graphics}
\usepackage{times}
\usepackage{graphics}
\usepackage{epsfig}
\usepackage{amsfonts}
\usepackage{graphicx}
\usepackage{makecell}
\usepackage{indentfirst}
\usepackage{amssymb}
\usepackage{xcolor}
\DeclareMathOperator*{\argmax}{argmax}
\DeclareMathOperator*{\Softmax}{Softmax}
\usepackage{multirow}
\usepackage[ruled,vlined,linesnumbered]{algorithm2e}
\usepackage[]{hyperref}
\modulolinenumbers[5]

\journal{Journal of Information Processing and Management}

\begin{document}

\begin{frontmatter}

\title{Deep Ranking Based Cost-sensitive Multi-label Learning for Distant Supervision Relation Extraction}

\author{Hai Ye}
\address{School of Computer Science and Engineering, Beihang University, Beijing, China}
\ead{hye.me@outlook.com}

\author{Zhunchen Luo\corref{mycorrespondingauthor}}
\cortext[mycorrespondingauthor]{corresponding author}
\address{Information Research Center of Military Science, Beijing, China}
\address{PLA Academy of Military Science, Beijing, China}
\ead{zhunchenluo@gmail.com}




\begin{abstract}
Knowledge base provides a potential way to improve the intelligence of information retrieval (IR) systems, for that knowledge base has numerous relations between entities which can help the IR systems to conduct inference from one entity to another entity. Relation extraction is one of the fundamental techniques to construct a knowledge base. Distant supervision is a semi-supervised learning method for relation extraction which learns with labeled and unlabeled data. However, this approach suffers the problem of \emph{relation overlapping} in which one entity tuple may have multiple relation facts. We believe that relation types can have latent connections, which we call \emph{class ties}, and can be exploited to enhance relation extraction. However, this property between relation classes has not been fully explored before. In this paper, to exploit class ties between relations to improve relation extraction, we propose a general ranking based multi-label learning framework combined with convolutional neural networks, in which ranking based loss functions with regularization technique are introduced to learn the latent connections between relations. Furthermore, to deal with the problem of \emph{class imbalance} in distant supervision relation extraction, we further adopt cost-sensitive learning to rescale the costs from the positive and negative labels. 
Extensive experiments on a widely used dataset show the effectiveness of our model to exploit class ties and to relieve class imbalance problem. 
\end{abstract}


\end{frontmatter}


\section{Introduction}
\begin{figure*}
\centering\includegraphics[width = 10cm]{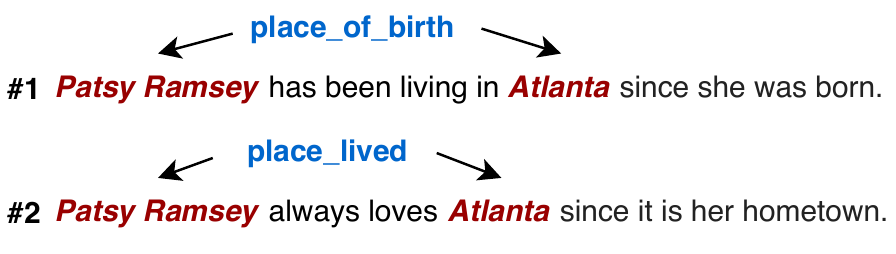}
\caption{Training instances generated by freebase. The entity tuple is (\emph{Patsy Ramsey, Atlanta}) and its two relation facts are \emph{palce\_of\_birth} and \emph{place\_lived}.}
\label{training_instance}
\end{figure*}

{R}{elation} extraction (RE) 
aims to classify the relations~(or called relation facts) between two given named entities from natural-language text. 
Fig.~\ref{training_instance} shows two sentences with the same entity tuple but two different relation facts. RE is to accurately extract the corresponding relation facts~(\emph{place\_of\_birth}, \emph{place\_lived}) for the entity tuple~(\emph{Patsy Ramsey, Atlanta}) based on the contexts of sentences. 
Supervised-learning methods require numerous labeled data to work well. 
With the rapid growth of volume of relation types, traditional methods can not keep up with the step for the limitation of labeled data. In order to narrow down the gap of data sparsity, \cite{2009} proposes \emph{distant supervision (DS)} for relation extraction, which automatically generates training data by aligning a knowledge facts database (ie. Freebase~\cite{freebase}) to texts. For a fact~(e.g. entity tuple with a relation type) from the knowledge base, the sentences containing the entity tuple in the fact are regarded as the training data. 

\emph{Class ties} mean the connections~(relatedness) between relations types for relation extraction. 
In general, we conclude that class ties can have two categories: weak class ties and strong class ties. Weak class ties mainly involve the co-occurrence of relations such as \emph{place\_of\_birth} and \emph{place\_lived}, \emph{CEO\_of} and \emph{founder\_of}. Besides, strong class ties mean that relations have latent logical entailments. Take the two relations of \emph{capital\_of} and \emph{city\_of} for example, if one entity tuple has the relation of \emph{capital\_of}, it must express the relation fact of \emph{city\_of}, because the two relations have the entailment of \emph{capital\_of} $\Rightarrow$ \emph{city\_of}. Obviously the opposite induction is not correct.
Further take the following sentence of 
\begin{quote}
\emph{Jonbenet told me that her mother $\text{[Patsy Ramsey]}_{e_1}$ never left $\text{[Atlanta]}_{e_2}$ since she was born.}
\end{quote}
for example. This sentence expresses two relation facts which are \emph{place\_of\_birth} and \emph{place\_lived}.
However, the word ``born" is a strong bias to extract \emph{place\_of\_birth}, so it may not be easy to predict the relation of \emph{place\_lived}, but extracting \emph{place\_of\_birth} will provide evidence for prediction of \emph{place\_lived} by incorporating the weak ties between the two relations,

Exploiting class ties is necessary for DS based relation extraction. In DS scenario, there is a challenge that one entity tuple can have multiple relation facts which is called \emph{relation overlapping}~\cite{2011,2012}, as shown in Fig.~\ref{training_instance}. However, the relations of one entity tuple can have class ties mentioned above which can be leveraged to enhance relation extraction, for that it narrows down potential searching spaces and reduces uncertainties between relations when predicting unknown relations, such that if one pair of entities has \emph{CEO\_of} relation, it will contain \emph{founder\_of} relation with high possibility. 

To exploit class ties between relations, we propose to make joint extraction by considering \emph{pairwise} connections between positive and negative labels inspired by~\cite{furnkranz2008multilabel,zhang2006multilabel}. As the example for one entity tuple with two different relation types shown in Fig.~\ref{training_instance}, by extracting the two relations jointly, we can maintain the \emph{class ties} (co-occurrence) of them and the class ties can be learned by potential models, which  can be leveraged to extract instances with unknown relations. 
We introduce a ranking based multi-label learning framework to make joint extraction, to learn to rank the prediction probability for positive relations higher than negative ones. We design ranking based loss functions for multi-label learning. Furthermore, inspired by~\cite{zhouMIML,MIML2005}, we add a regularization term to the loss functions to better learn the relatedness between relation facts, and we only regularize the positive relation types ignoring the relation of NR (does not express any relation) based on the assumption that the connections between relations are only in positive relations but not in NR~(see Sec.~\ref{Learning Class Ties via Ranking based Multi-label Learning with Regularization}).

\begin{figure}
\centering\includegraphics[width = 7cm]{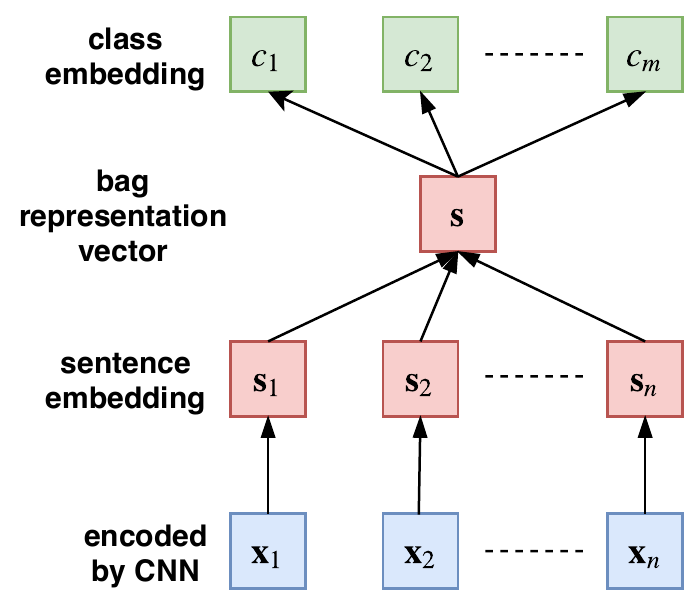}
\caption{The main architecture of our model. The features of sentences are encoded by CNN model, and then the sentence embeddings are aggregated, finally the bag representation is used to make joint extraction.}
\label{archi}
\end{figure}

Besides, class imbalance is the another severe problem which can not be ignored for distant supervision relation extraction. We find that around $\mathrm{70}\%$ training data express NR relation type and even more than $\mathrm{90}\%$ in test set, so samples with NR type count a much higher proportion comparing to the positive samples~(not categorized as NR). This problem will severely affect the model training, causing the model easily to classify the samples to have the NR relation type~\cite{japkowicz2002class}. To overcome this problem, based on the ranking loss functions, we further adopt cost-sensitive learning to rescale the costs from the positive and negative labels, by increasing the losses for positive labels and penalizing losses from NR type~(detailed in Sec.~\ref{Ranking based Cost-sensitive Multi-label Learning}).

Furthermore, combining information across sentences will be more appropriate for joint extraction which provides more information from other sentences to extract each relation (\cite{hao,lin2016}). In Fig.~\ref{training_instance}, sentence \#1 is the evidence for \emph{place\_of\_birth}, but it also expresses the meaning of ``living in someplace'', so it can be aggregated with sentence \#2 to extract \emph{place\_lived}. Meanwhile, the word of ``hometown'' in sentence \#2 can provide evidence for \emph{place\_of\_birth} which should be combined with sentence \#1 to extract \emph{place\_of\_birth}.

In this work, we propose a unified model that integrates ranking based cost-sensitive multi-label learning with convolutional neural network~(CNN) to exploit class ties between relations and further relieve the class imbalance problem. 
Inspired by the effectiveness of deep learning for modeling sentence features~\cite{deeplearning2015}, we use CNN to encode sentences. Similar to~\cite{lin2016,rank2015}, we use class embeddings to represent relation classes. 
The whole model architecture is presented in Fig.~\ref{archi}. 
We first use CNN to embed sentences, 
then we introduce two variant methods to combine the embedded sentences into one bag representation vector aiming to aggregate information across sentences, 
after that we measure the similarity between the bag representation and relation class in real-valued space. Finally, we use the ranking loss functions to learn to make joint extraction over multiple relation types.

Our experimental results on dataset of~\cite{2010} are evident that: (1) Our model is much more effective than the baselines; (2) Leveraging class ties will enhance relation extraction and our model is efficient to learn class ties by joint extraction; (3) A much better model can be trained after relieving class imbalance from NR.

Our contributions in this paper can be encapsulated as follows:

$\bullet$ We propose to leverage class ties to enhance relation extraction. Combined with CNN,  an effective deep ranking based multi-label learning model with regularization technique is introduced to exploit class ties. 

$\bullet$ We adopt the cost-sensitive learning to relieve the class imbalance problem and experimental results show the effectiveness of our method.


\section{Related Work}
\subsection{Relation Extraction}
Previous methods on relation extraction can mainly be summarized as supervision based and distant supervision based. Supervision based methods needs much labeled data to work well which can not keep up with the rapid growth of relation types. To overcome the problem of data sparsity for supervision based methods, distant supervision relation extraction has been proposed by~\cite{2009}. However, DS based relation extraction suffers the two problems of \emph{wrong labelling problem} and \emph{overlapping problem}, in which the former means that sentences containing certain entities actually do not express the relation type of the entities indicated or even do not express any relations and the latter mean that one entity tuple may have multiple relation types. 
To solve the problem of wrong labelling, \cite{2010} introduces multi-instance learning for relation extraction in which the mentions of one certain entity tuple are merged as one bag and make the model to extract relations on mention bags, however this method can not deal with the relation overlapping problem. Afterwards, \cite{2011} and \cite{2012} introduce the framework of multi-instance multi-label learning to jointly overcome the two problems and improve the performance significantly. Though they also propose to make joint extraction of relations, they only use information from single sentence losing information from other sentences. \cite{Global} tries to use \emph{Markov logic} model to capture consistency between relation labels, on the contrary, our model leverages deep ranking to learn class ties automatically. 

Recent years, deep learning has achieved remarkable success in computer vision and natural language processing~\cite{deeplearning2015}. 
Deep learning has been applied to automatically learn the features of sentences (\cite{zeng2014, Yu2014, rank2015,lin2016,DBLP:conf/pakdd/YeYLC17,DBLP:conf/emnlp/YeW18,DBLP:journals/corr/abs-1802-08504,DBLP:conf/coling/JiangYLCM18}). In supervision relation extraction, \cite{zeng2014} applies convolutional neural networks to model sentences and import position feature for RE, which obtains significant gains in RE performance. Afterwards, \cite{Yu2014, rank2015, lin2016} further introduce more advanced deep learning models for RE. 
In distant supervision relation extraction, \cite{zeng2015} proposes a piecewise convolutional neural network with multi-instance learning for DS based relation extraction, which improves the precision and recall significantly. Afterwards, \cite{lin2016} introduces the attention mechanism (\cite{attention1,attention2}) to merge the sentence features aiming to construct better bag representations. \cite{lin2017} further proposes a multi-lingual neural relation extraction framework considering the information consistency and complementarity among cross-lingual texts. However, the two deep learning based models only make separated extraction thus can not model class ties between relations. Recently, \cite{zeng2016incorporating} proposes to incorporate relation paths for distant supervision relation extraction and \cite{ji2017distant} introduces to use the description of entities to enhance distant supervision relation extraction. \cite{chen2018encoding} proposes a joint inference approach by encoding implicit relation requirements for relation extraction. Joint learning is also applied to jointly study two related tasks~\cite{DBLP:journals/corr/abs-1906-00575}. 
Besides, a lot of works have been proposed in recent times to solve the wrong labelling problem. \cite{bingfeng2017} proposes to model the noise caused by wrong labelling problem and show that dynamic transition matrix can effectively characterize the noises. \cite{qin2018dsgan, han2018denoising} propose to use adversarial learning~\cite{goodfellow2014generative} to solve the wrong labelling problem. Instead, \cite{DBLP:conf/aaai/FengHZYZ18, DBLP:conf/acl/WangXQ18a} adopt reinforcement learning to learn to select high-quality data for training. \cite{DBLP:conf/emnlp/LiuWCS17} dynamically corrects the wrong labeled data during training by exploiting semantic information from labeled entity pairs. \cite{DBLP:conf/emnlp/LiuZZJ18} transfers the priori knowledge learned from relevant entity classification task to make the model robust to noisy data.

\subsection{Deep Learning to Rank}
Learning to rank (LTR) is an important technique in information retrieval (IR)~\cite{liu2009}. The methods to train a LTR model include pointwise, pairwise and listwise. We apply pairwise LTR in our paper. Deep learning to rank has been widely used in many problems to serve as a classification model. In image retrieval, \cite{zhao2015deep} applies deep semantic ranking for multi-label image retrieval. In text matching, \cite{severyn2015learning} adopts learning to rank combined with deep CNN for short text pairs matching. In traditional supervised relation extraction, \cite{rank2015} designs a pairwise loss function based on CNN for single label relation extraction. Based on the advantage of deep learning to rank, we propose pairwise learning to rank (LTR)~\cite{liu2009} combined with CNN in our model aiming to jointly extract multiple relations. 

\subsection{Cost-sensitive Learning}
Cost-sensitive learning is one of the techniques for class imbalance problem, which assigns higher wrong classification costs to classes with small proportion. For example, \cite{shen2015deepcontour} proposes a regularized softmax to deal with the imbalanced edge label classification. \cite{khan2015cost} adopts cost-sensitive learning to learn deep feature representations from imbalanced data. Another approach to relieve class imbalance problem is re-sampling \cite{huang2016learning,imbalance} including over-sampling and under-sampling, which aims to balance the distributions of data in different labels. 

This paper is the extension of~\cite{yehaiacl2017}. Compared to original work in~\cite{yehaiacl2017}, this paper has several improvements: 

 \textbf{Methods}: (a) We further fully consider the class imbalance problem. We propose a novel ranking based cost-sensitive loss function combined with multi-label learning. (b) To better learn class ties between relations, we further introduce a regularization term to ranking loss functions.

\textbf{Experiments}: (a) We further do experiments to analyze the effectiveness of our novel cost-sensitive ranking loss functions. (b) The evaluation experiments on the effectiveness of regularization have further be conducted. 

\textbf{Content}: (a) We rewrite the description of our methods from the view of multi-label learning and cost-sensitive learning to gain more theoretical justification improvement.

\section{Methodology}
We introduce our methods in this section. Firstly, we describe the widely used CNN architecture for sentence encoding. Then we discuss the ranking based multi-label learning framework with regularization technique. After that, we introduce the proposed cost-sensitive learning to overcome the NR effects for model training.

\subsection{Notation}
We define the relation classes as $\mathcal{L} = \{1,2,\cdots,C\}$, entity tuples as $\mathcal{T}=\{t_i\}_{i=1}^M$ and mentions\footnote{The sentence containing one certain entity is called mention.} as $\mathcal{X} = \{x_i\}_{i=1}^N$. Dataset is constructed as follows: for entity tuple $t_i \in \mathcal{T}$ and its relation class set $L_i \subseteq \mathcal{L}$, we collect all the mentions $X_i$ that contain $t_i$, the dataset we use is $\mathcal{D} = \{(t_i, L_i, X_i)\}_{i=1}^H$.
Given a data $(t_k, L_k, X_k) \in \mathcal{D}$, the sentence embeddings of $X_k$ encoded by CNN are defined as $S_k = \{{s_i}\}_{i=1}^{|X_k|}$ and 
we use class embeddings $W \in R^{|\mathcal{L}| \times d}$ to represent the relation classes, which will be learned in model training.

\subsection{CNN for sentence embedding}
We take the effective piecewise CNN architecture adopted from~\cite{zeng2015,lin2016} to encode sentence and we will briefly introduce PCNN in this section. More details of PCNN can be obtained from previous work.

\subsubsection{Words Representations}\

{$\bullet$ \textbf{Word Embedding}  Given a word embedding matrix $V \in \mathbb{R}^{l^w \times d^1}$ where $l^w$ is the size of word dictionary and $d^1$ is the dimension of word embedding, the words of a mention $x = \{w_1, w_2, \cdots, w_n\}$ will be represented by real-valued vectors from $V$.}

{$\bullet$ \textbf{Position Embedding} The position embedding of a word measures the distance from the word to entities in a mention.
We add position embeddings into words representations by appending position embedding to word embedding for every word. Given a position embedding matrix $P \in \mathbb{R}^{l^p \times d^2}$ where $l^p$ is the number of distances and $d^2$ is the dimension of position embeddings, the dimension of words representations becomes $d^w = d^1 + d^2 \times 2$.}

\subsubsection{Convolution, Piecewise max-pooling}\

After transforming words in $x$ to real-valued vectors, we get the sentence $q \in {\mathbb{R}}^{n \times d^w}$. The set of kernels $K$ is $\{ {K_i}\}_{i=1}^{d^s}$ where $d^s$ is the number of kernels. Define the window size as $d^{win}$ and given one kernel $K_k \in {\mathbb{R}}^{d^{win} \times d^w}$, the convolution operation is defined as follows:
\begin{equation}
m_{[i]} = q_{[i:i+d^{win}-1]} \odot K_k + b_{[k]}
\end{equation}
where $m$ is the vector after conducting convolution along $q$ for $n - d^{win} + 1$ times and $b \in {\mathbb{R}}^{d^s}$ is the bias vector.
For these vectors whose indexes out of range of $[1,n]$, we replace them with zero vectors.



By piecewise max-pooling, when pooling, the sentence is divided into three parts: $m_{[p_0:p_1]}$, $m_{[p_1:p_2]}$ and $m_{[p_2:p_3]}$ ($p_1$ and $p_2$ are the positions of entities, $p_0$ is the beginning of sentence and $p_3$ is the end of sentence). This piecewise max-pooling is defined as follows:
\begin{equation}
z_{[j]} = max(m_{[p_{j-1}:p_j]})
\end{equation}
where $z \in \mathbb{R}^3$ is the result of mention $x$ processed by kernel $K_k$; $1 \le j \le 3$. 
Given the set of kernels ${K}$, following the above steps, the mention $x$ can be embedded to $o$ where $o \in R^{d^s * 3}$.
\subsubsection{Non-Linear Layer, Regularization}
To learn high-level features of mentions, we apply a non-linear layer after pooling layer. After that, a dropout layer is applied to prevent over-fitting. We define the final fixed sentence representation as $r \in {\mathbb{R}}^{d^f}$ ($d^f = d^s * 3$).
\begin{equation}
s = g(o) \circ h
\end{equation}
where $g(\cdot)$ is a non-linear function and we use $tanh(\cdot)$ in this paper; $h$ is a Bernoulli random vector with probability p to be $1$.

\subsection{Combine Information across Sentences}

We propose two options to combine sentences to provide enough information for multi-label learning.

\noindent{$\bullet$ \textbf{AVE} \ \ The first option is average method. This method regards all the sentences equally and directly average the values in all dimensions of sentence embedding. This \textbf{AVE} function is defined as follows:}
\begin{equation}
\label{func4}
r = \frac{1}{n}\sum_{s_i \in S_k} s_i
\end{equation}
where $n$ is the number of sentences and $r$ is the bag representation combining all sentence embeddings.
Because it weights the importance of sentences equally, this method may bring much noise data from two aspects: (1) the wrong labelling data; (2) irrelated mentions for one relation class, for all sentences containing the same entity tuple being combined together to construct the bag representation. 

\noindent{$\bullet$ \textbf{ATT} \ \ The second one is a sentence-level attention algorithm used by ~\cite{lin2016} to measure the importance of sentences aiming to relieve the wrong labelling problem. For every sentence, \textbf{ATT} will calculate a weight by comparing the sentence to one relation. We first calculate the similarity between one sentence embedding and relation class as follows:}
\begin{equation}
e_j = a \cdot W_{[c]} \cdot s_j
\label{func5}
\end{equation}
where $e_j$ is the similarity between sentence embedding $s_j$ and relation class $c$ and a is a bias factor. In this paper, we set $a$ as $0.5$. Then we apply $\Softmax$ to rescale $\mathbf{e}$ ($\mathbf{e} = \{e_i\}_{i=1}^{|X_k|}$) to $[0,1]$. We get the weight ${\alpha}_j$ for $s_j$ as follows:
\begin{equation}
{\alpha}_j = \frac{\exp(e_j)}{\sum_{e_i \in \mathbf{e}} \exp(e_i)}
\label{func6}
\end{equation}
so the function to merge $r$ with \textbf{ATT} is as follows:
\begin{equation}
r = \sum_{i=1}^{|X_k|} \alpha_i \cdot s_i
\label{func7}
\end{equation}

\subsection{Learning Class Ties via Ranking based Multi-label Learning with Regularization}\label{Learning Class Ties via Ranking based Multi-label Learning with Regularization}
Firstly, we have to present the score function to measure the similarity between bag representation $r$ and relation $c$.

\noindent{$\bullet$ \textbf{Score Function} \ \ We use dot function to produce score for $r$ to be predicted as relation $c$. The score function is as follows:}
\begin{equation}
\mathcal{F}(r,c) = W_{[c]}\cdot r
\end{equation}

There are other options for score function. In~\cite{MultiATT}, they propose a margin based loss function that measures the similarity between $r$ and $W_{[c]}$ by distance. Because score function is not an important issue in our model, we adopt dot function, also used by~\cite{rank2015} and~\cite{lin2016}, as our score function.

Now we start to introduce the ranking loss functions.

Pairwise ranking aims to learn the score function $\mathcal{F}(r,c)$ that ranks positive classes higher than negative ones. This goal can be summarized as follows:
\begin{equation}
 \forall c^+ \in L_k, \forall c^- \in \mathcal{L}-L_k : {\mathcal{F}}(r,c^+) > {\mathcal{F}}(r,c^-) + \beta
 \label{frank}
\end{equation}
where $\beta$ is a margin factor which controls the minimum margin between the positive scores and negative scores. Inspired by~\cite{rank2015}, given $c^+$ and $c^-$, we adopt the following function to learn the score function:
\begin{gather}
\nonumber \mathcal{H}(c^+,c^-,r) = \ln(1+\exp(\rho[0, \sigma^+ - \mathcal{F}(r, c^+)]))  \\
 + \ln(1 + \exp(\rho[0, \sigma^- + \mathcal{F}(r,c^-)]))
 \label{frank1}
\end{gather}
where $[0, \cdot] = \max(0, \cdot)$, $\rho$ is the rescale factor, $\sigma^+$ is positive margin and $\sigma^-$ is negative margin.
This loss function is designed to rank positive classes higher than negative ones controlled by the margin of $\sigma^+ - \sigma^-$.
In reality, $\mathcal{F}(r,c^+)$ will be higher than $\sigma^+$ and $\mathcal{F}(r,c^-)$ will be lower than $\sigma^-$. In our work, we set $\rho$ as $2$, $\sigma^+$ as $2.5$ and $\sigma^-$ as $0.5$ adopted from ~\cite{rank2015}. To simplify the loss functions given in the followings, we use $\rho[0, \sigma^+ - \mathcal{F}(r, c^+)]$ to replace the first term in $\mathcal{H}$ and use $\rho[0, \sigma^- + \mathcal{F}(r, c^-)]$ to replace the second term.

To model the class ties~(co-occurrence) of the labels, we have the assumption that the positive labels have the same class ties and are connected with each other. Out of this assumption, we have two mechanisms to learn the class ties, which are making joint extraction of relations and explicitly modeling the connections by regularizing the learning of positive labels. In the followings, we will first introduce the loss functions for multi-label learning extended from Eq.~\ref{frank1}; then we discuss the regularization term. 

To learn class ties between relations, we firstly extend the Eq.~\ref{frank1} to make multi-label learning. Followings are the proposed ranking based loss functions: 


\noindent{$\bullet$ \textbf{with AVE (Variant-1)} \ \ We define the margin-based loss function with option of \textbf{AVE} to aggregate sentences as follows:}
\begin{gather}
\nonumber G_{[\mathrm{ave}]} = \sum_{c^+ \in L_k} \rho[0, \sigma^+ - \mathcal{F}(r,c^+)] \\+ \rho|L_k|[0, \sigma^- + \mathcal{F}(r,c^-)]
\label{func9}
\end{gather}
Similar to~\cite{Weston2011} and~\cite{rank2015}, we update one negative class at every training round but to balance the loss between positive classes and negative ones, we multiply $|L_k|$ before the right term in Eq.~\ref{func9} to expand the negative loss. We apply mini-batch based stochastic gradient descent (SGD) to minimize the loss function. The negative class is chosen as the one with highest score among all negative classes~\cite{rank2015}, i.e.:
\begin{equation}
c^- = \mathop{\argmax}_{c \in \mathcal{L}-L_k} \mathcal{F}(r,c)
\end{equation}

\noindent{$\bullet$ \textbf{with ATT (Variant-2)} \ \  Now we define the loss function for the option of \textbf{ATT} to combine sentences as follows:}
\begin{gather}
\nonumber G_{[\mathrm{att}]} = \sum_{c^+ \in L_k} \big \{ \rho[0, \sigma^+ - \mathcal{F}(r^{c^+},c^+)] \\ + \rho [0, \sigma^- + \mathcal{F} (r^{c^+}, c^-)] \big \}
\label{func11}
\end{gather}
where $r^{c}$ means the attention weighted representation $r$ where attention weights are merged by comparing sentence embeddings with relation class $c$ and $c^-$ is chosen by the following function:
\begin{equation}
c^- = \mathop{\argmax}_{c \in \mathcal{L}-L_k} \mathcal{F}(r^{c^+},c)
\label{func13}
\end{equation}
which means we update one negative class in every training round.
We keep the values of $\rho$, $\sigma^+$ and $\sigma^-$ same as values in Eq.~\ref{func9}. 
In Eq.~\ref{func11}, for every $c^+ \in L_k$, we need to sample $c^- \in \mathcal{L}-L_k$ according to Eq.~\ref{func13}, so different from Eq.~\ref{func9}, we do not extend the negative loss by multiplying $|L_k|$. 

According to this loss function, we can see that: for each class $c^+ \in L_k$, it will capture the most related information from sentences to merge $r^{c^+}$, then rank $\mathcal{F}(r^{c^+},c^+)$ higher than all negative scores which each is $\mathcal{F}(r^{c^+},c^-)$ ($c^- \in \mathcal{L} - L_k$). We use the same update algorithm to minimize this loss.

Based on the assumption that all positive labels have the same class ties, making joint extraction of the relations can capture the co-occurrence of the labels. If the relations for the same entity pair usually appear together, then extracting them jointly can learn the statistical property of their co-appearance.

\noindent $\bullet$ \textbf{Regularization} To learn the class ties between relations, we have proposed the ranking based loss functions above. Inspired by~\cite{zhouMIML,MIML2005}, we further capture the relation connections by adding an extra regularization term to the loss functions. We only consider the relatedness between positive labels ignoring NR. The relatedness is measured by the mean function $W_{ave}$:
\begin{equation}
W_{ave} = \frac{1}{T} \sum_{c \in \mathcal{L}-c_{\mathrm{NR}}}W_{[c]}
\label{mean}
\end{equation}
where $T = |\mathcal{L}-c_{\mathrm{NR}}|$. $W_{ave}$ is the center of the labels, and we hope the positive labels can be close to the center which can be measured by:
\begin{equation}
\frac{1}{T} \sum_{c \in \mathcal{L} - c_{NR}} \Vert W_{[c]} - W_{ave} \Vert_{2} 
\label{mean1}
\end{equation}
Following~\cite{zhouMIML}, to model the class ties we need to minimize the loss function as follows: 
\begin{equation}
\Theta(W) = \epsilon \Vert W_{ave} \Vert_{2} + \eta\frac{1}{T}\sum_{\mathcal{L}-c_{NR}}\Vert W_{[c]} - W_{ave} \Vert_2
\label{regu1}
\end{equation}
where $\epsilon$ and $\eta$ are hyper-parameters. Eq.~\ref{regu1} is designed based on the consideration that the labels in which class ties exist should be clustered together and should be close to the center of these labels. 
According to Eq.~\ref{mean}, Eq.~\ref{mean1} can be re-written as:
\begin{equation}
- \Vert W_{ave} \Vert_2 + \frac{1}{T} \sum_{c \in \mathcal{L} - c_{NR}} \Vert W_{[c]} \Vert_2  
\label{mean2}
\end{equation}
By merging Func.~\ref{mean2} into Eq.~\ref{regu1}, we have the our final regularization term: 
\begin{equation}
\Theta(W) =  \epsilon \Vert W_{ave} \Vert_{2} + \eta\frac{1}{T}\sum_{c \in \mathcal{L}-c_{\mathrm{NR}}} \Vert W_{[c]} \Vert_{2} 
\end{equation}
In this paper, we set $\eta$ as $10^{-3}$ and $\epsilon$ is set as $10^{-6}$.
 

\begin{table}[]
\caption{The proportions of NR samples from Riedel's dataset.}
\centering
\begin{tabular}{l|cc}
\hline
\textbf{Pro. (\%)} & \textbf{Training} & \textbf{Test} \\
\hline
\textbf{Riedel}    & 72.52             & 96.26        \\
\hline
\end{tabular}
\label{NR1}
\end{table}

\subsection{Ranking based Cost-sensitive Multi-label Learning}\label{Ranking based Cost-sensitive Multi-label Learning}
In relation extraction, the dataset will always contain certain negative samples which do not express any relation types and are classified as NR type~(no relation).
Table~\ref{NR1} presents the proportion of NR samples in the 
dataset from~\cite{2010}, which shows that the almost data is about NR. Data imbalance will severely affect the model training and cause the model only sensitive to classes with high proportion~\cite{imbalance}, causing a positive sample to be classified as NR. In order to relieve this problem, we adopt cost-sensitive learning to construct the loss function. Based on $G_{[\mathrm{att}]}$, the cost-sensitive loss function which is \textbf{Variant-3} is as follows:
\begin{gather} 
\nonumber G_{[\mathrm{cost\_att]}} = \sum_{c^* \in L_k}\big \{ g(c^*) \big (\rho[0, \sigma^+ - \mathcal{F}(r^{c^*}, c^*)] \big) \\ 
\nonumber + \rho[0, \sigma^- + \mathcal{F}(r^{c^*}, c^-)] \\ 
\nonumber + \sum_{c^+ \in L_k - c^*}  \gamma \rho[0, \sigma^+ - \mathcal{F}(r^{c^*}, c^+)]  \\ 
 + \gamma \mathbf{1}(c^* \ne c_{\mathrm{NR}})\rho[0, \sigma^- + \mathcal{F}(r^{c^*}, c_{\mathrm{NR}})] \big \}
\end{gather}
where $g(c^*) = \textbf{1}(c = c_{\mathrm{NR}})\lambda + \textbf{1}(c \ne c_{\mathrm{NR}})1$; $\textbf{1}(\cdot)$ is an indicate function. 
Similar to Eq.~\ref{func13}, we select $c^-$ as follows:
\begin{equation}
c^- = \mathop{\argmax}_{c \in \mathcal{L}-L_k} \mathcal{F}(r^{c^*},c)
\label{func15}
\end{equation}

Because NR counts a high proportion in the training set, without controlling, the model will receive large costs from NR. In order to relieve the effects from NR, we penalize the losses from NR. Specifically, we have two strategies to do that. We adopt two hyper-parameters which are $\lambda$~($\lambda < 1$) and $\gamma$ to penalize the losses from NR. If $c^* \in L_k$ is a positive label, to balance the costs between the positive labels and the NR label, we further add the costs from the left positive relations $c^+ \in L_k - c^*$  and at the same time, the extra cost from NR is calculated. 
The default value of $\gamma$ is $1$ and if $\gamma$ is small enough, this loss function will be similar to loss Eq.~\ref{func11}. Based on the experimental results, we find that the best results are achieved when $\lambda$ is set to $0$, so we set $\lambda$ as $0$ in this paper. How the $\lambda$ and $\gamma$ affect model performance is discussed in Sec.~\ref{ImpactofCost-sensitiveLearning} and Sec.~\ref{ImpactOfNR}. We also add the regularization term $\Theta(W)$ to $G_{[\mathrm{cost\_att}]}$ to better capture the class ties between relations.

We give out the pseudocode of merging $G_{[\mathrm{cost\_att}]}$ in algorithm $1$.

\IncMargin{0.5em}
\begin{algorithm}
\SetKwData{Left}{left}\SetKwData{This}{this}\SetKwData{Up}{up}
\SetKwFunction{Union}{Union}\SetKwFunction{FindCompress}{FindCompress}
\SetKwInOut{Input}{input}\SetKwInOut{Output}{output}
\Input{$\mathcal{L}$, $(t_k, L_k, X_k)$ and $S_k$;}
\Output{$G_{[\mathrm{cost\_att]}}$;}
$G_{[\mathrm{cost\_att}]} \gets 0$\;
\For{$c^* \in L_k$}{\label{forins}
Merge representation $r^{c^*}$ by Eq.~\ref{func5}, \ref{func6}, \ref{func7}\;

$G_{[\mathrm{cost\_att}]} \gets g(c^*)(\rho[0, \sigma^+ - \mathcal{F}(r^{c^*},c^*)])$\;

$c^- \gets \argmax_{c \in \mathcal{L}-L_k} \mathcal{F}(r^{c^*}, c)$\;

$G_{[\mathrm{cost\_att}]} \gets G_{[\mathrm{cost\_att}]} + \rho [0, \sigma^- + \mathcal{F}(r^{c^*}, c^-)]$\;

\For{$c^+ \in L_k-c^*$}{

$G_{[\mathrm{cost\_att}]} \gets G_{[\mathrm{cost\_att}]} + \gamma \rho[0, \sigma^+ - \mathcal{F}(r^{c^*},c^+)]$\;

}
$G_{[\mathrm{cost\_att}]} \gets G_{[\mathrm{cost\_att}]} 
 + \gamma \mathbf{1}(c^* \ne c_{\mathrm{NR}}) \rho[0, \sigma^- + \mathcal{F}(r^{c^*}, c_{\mathrm{NR}})] $\;
}
\caption{Ranking based Cost-sensitive Multi-label Learning}
return $G_{[\mathrm{cost\_att}]}$\;
\end{algorithm}\DecMargin{0.5em}

\section{Experiments}
In this section, we conduct two sets of experiments, in which the first one is for comparing our method with the baselines and the second one is used to evaluate our model. Without the special statement, we will adhere to the methods and settings mentioned above to conduct the following experiments.
\subsection{Dataset and Evaluation Criteria}
\noindent{\textbf{Dataset.}} \ \ We conduct our experiments on a widely used dataset, developed by~\cite{2010} and has been used by~\cite{2011,2012,zeng2015,lin2016}. The dataset aligns Freebase relation facts with the New York Times corpus, in which training mentions are from 2005-2006 corpus and test mentions from 2007. The training set contains 522,611 sentences, 281,270 entity pairs and 18,252 relation facts. In test set, there are 172,448 sentences, 96,678 entity pairs and 1,950 relation facts. In all, there are 53 relation labels including the NR relation. Following~\cite{2009}, we adopt held-out evaluation framework in all experiments. We use all training dataset to train our model and then test the trained model on test dataset to compare the predicted relations to gold relations.

\noindent{\textbf{Evaluation Criteria.}} \ \ To evaluate the model performance, we draw the precision/recall (P/R) curves and precision@N (P@N) is reported to illustrate the model performance. For the metric of P/R curve, the bigger of the area contained under the curve, the better of the model performance. 
\subsection{Experimental Settings}
\noindent{\textbf{Word Embeddings.} \  \
We adopt the trained word embeddings from~\cite{lin2016}. 
Similar to~\cite{lin2016}, we keep the words that appear more than $100$ times to construct word dictionary and use ``UNK'' to represent the other ones.}

\noindent{\textbf{Hyper-parameter Settings.} \ \
Three-fold validation on the training dataset is adopted to tune the parameters following~\cite{2012}. We select word embedding size from $\{50,100,150,200,250,300\}$. Batch size is tuned from $\{80, 160, 320, 640\}$. We determine learning rate among $\{0.01,0.02,0.03,0.04\}$. The window size of convolution is tuned from $\{1, 3 ,5\}$. We keep other hyper-parameters same as~\cite{zeng2015}: the number of kernels is $230$, position embedding size is $5$ and dropout rate is $0.5$. Table~\ref{parameter} shows the detailed parameter settings.} 

\begin{table}
  \centering
  \caption{Hyper-parameter settings.}
  \begin{tabular}{l|l|r}
  \hline
  \textbf{Parameter Name} & \textbf{Symbol} & \textbf{Value} \\ \hline
  Window size & $d^{win}$& $3$ \\
  Sentence. emb. dim. & $d^f$ & $690$ \\
  Word. emb. dim. & $d^1$ & $50$ \\
  Position. emb. dim. & $d^2$ & $5$ \\
  Batch size & $\mathcal{B}$ & $160$ \\
  Learning rate & $\mu$ & $0.03$ \\
  Dropout pos. & $p$ & $0.5$ \\ \hline 
  \end{tabular}
  \label{parameter}
\end{table}

\subsection{Comparisons with Baselines}
\noindent{\textbf{Baseline.}} \ \
We compare our model with the following baselines:

$\bullet$ \textbf{Mintz}~\cite{2009} is the first original model which incorporates distant supervision for relation extraction.

$\bullet$ \textbf{MultiR}~\cite{2011} is the multi-instance learning based graphical model which aims to address overlapping relation problem.

$\bullet$ \textbf{MIML}~\cite{2012} is a multi-instance multi-label framework which jointly considers the wrong labelling problem and overlapping problem.

$\bullet$ \textbf{PCNN+ATT}~\cite{lin2016} is the previous state-of-the-art model in dataset of~\cite{2010} which applies sentence-level attention to relieve the wrong labelling problem in DS based relation extraction. This model applies piece-wise convolutional neural network~\cite{zeng2015} to model sentences.

Besides comparing to the above methods, we also compare our variant models represented by \textbf{Rank+AVE}~(using loss function of $G_{[ave]}$), \textbf{Rank+ATT}~(using loss of $G_{[att]}$)and \textbf{Rank+Cost}~(using loss of $G_{[cost_att]}$). 


\begin{figure}
\centering\includegraphics[width = 10cm]{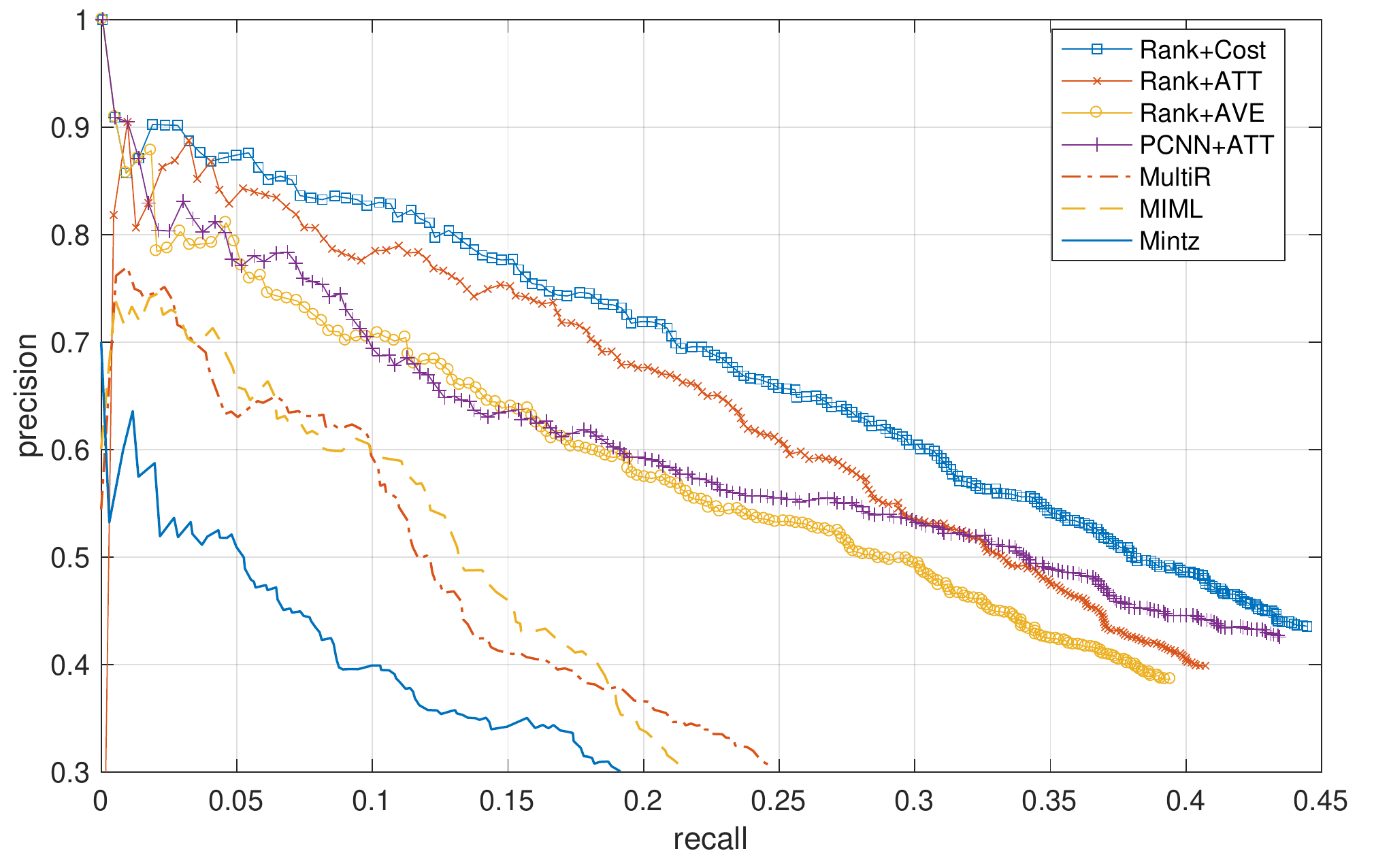}
\caption{\label{Fig:PC} Performance comparison of our model and the baselines. ``Rank+Cost" is using the loss function of $G_{[\mathrm{cost\_att}]}$, ``Rank+ATT" is using $G_{[\mathrm{att}]}$ and ``Rank+AVE" is using $G_{[\mathrm{ave}]}$.}
\end{figure}

\noindent{\textbf{Results and Discussion.} \ \
We compare our three variants of loss functions with the baselines and the results are shown in Fig.~\ref{Fig:PC}. From the results we can see that: 
\begin{itemize}
\item Rank+AVE (Variant-1) lags behind PCNN+ATT, whose reason may lie in that Rank+AVE does not use the attention mechanism to aggregate the information among the sentences, which brings much noise for encoding sentence contexts; 
\item After adopting the attention mechanism, Rank+ATT achieves much better performances comparing to Rank+AVE, and even better than PCNN+ATT;
\item Comparing PCNN+ATT and Rank+ATT, we can see that Rank+ATT is superior to PCNN+ATT, which comes from the strategy that we model the class ties into the relation extraction;
\item Our variant method of Rank+Cost achieves the best performance among all the baselines; by comparing to Rank+ATT, our cost-sensitive learning method can really work for relieving the negative effects from NR. 
\end{itemize}

\subsection{Impact of Class Ties}
In this section, we conduct experiments to reveal the effectiveness of our model to learn class ties with three variant loss functions mentioned above, and the impact of class ties for relation extraction. As mentioned above, we adopt two techniques to model the class ties: multi-label learning with ranking based loss functions and regularization term to better model class ties. In the followings, we will conduct experiments to reveal the two aspects for modeling class ties. We will adopt P/R curves and precisions@N ($100$, $200$, $\cdots$, $500$) to show the model performances.

\begin{figure}
\centering\includegraphics[width = 12cm]{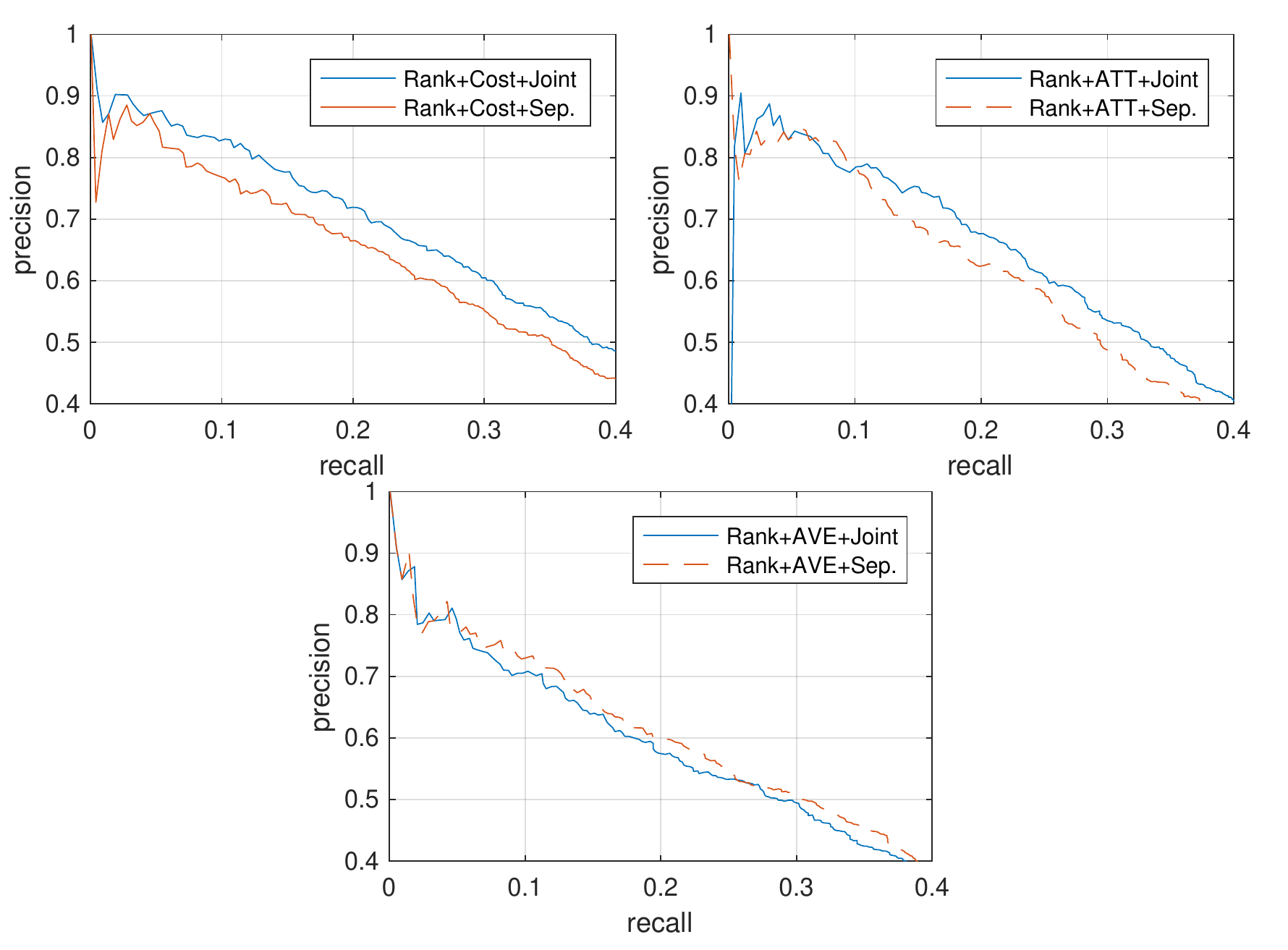}
\caption{Results for impact of ranking based loss function with methods of Rank + AVE, Rank + ATT and Rank + Cost}
\label{JointEx}
\end{figure}
$\bullet$ \textbf{Ranking based Loss Function.} The effectiveness of ranking loss functions to learn class ties lies in the joint extraction of relations to conduct multi-label leaning, so to reveal the impact of ranking loss function to learn class ties, we will compare the joint extraction with separated extraction. Regularization term is added to all variant models. To conduct the experiment of separated extraction, we divide the labels of entity tuple into single label and for one relation label we select the sentences expressing this relation to construct the bag, then we use the re-constructed dataset to train our model with our three variant loss functions. 


\begin{table}
 \caption{Precisions for top $100$, $200$, $300$, $400$, $500$ and average of them for impact of joint extraction and class ties.}\label{TabClassTies}
  \centering
  \begin{tabular}{p{2cm}|p{0.49cm}p{0.49cm}p{0.49cm}p{0.49cm}p{0.49cm}p{0.5cm}}
  \hline
  \textbf{P@N(\%)} & $\textbf{100}$ & $\textbf{200}$ & $\textbf{300}$ & $\textbf{400}$ & $\textbf{500}$ &\textbf{Ave.} \\ \hline

    \textbf{R.+AVE+J.}  & $79.1$  & $73.8$	&  $70.4$	& $66.0$	& $63.1$ & $70.5$\\ 
  \textbf{R.+AVE+S.}  & \textbf{80.2} & \textbf{74.9}	& \textbf{72.2}	& \textbf{67.8}	& \textbf{64.0} & \textbf{71.8}\\  \hline 
  \textbf{R.+ATT+J.} & \textbf{86.8} & $80.6$ & \textbf{78.4} & \textbf{75.2}	& \textbf{71.1} & \textbf{78.4} \\ 
  \textbf{R.+ATT+S.} & $82.4$	& \textbf{82.7}	& $75.3$	& $70.1$	& $66.2$ & 75.3\\ \hline
  \textbf{R.+ExATT+J.} & \textbf{86.8} & \textbf{83.2} & \textbf{81.1} & \textbf{76.7}	& \textbf{73.5} & \textbf{80.3} \\ 
  \textbf{R.+ExATT+S.} & $85.7$	& $78.5$	& $75.6$	& $72.4$	& $69.0$ & 76.3	 \\ \hline
  \end{tabular}
\end{table}

Experimental results are shown in Fig.~\ref{JointEx} and Table~\ref{TabClassTies}. From the results we can see that: (1) For Rank+ATT and Rank+Cost, joint extraction exhibits better performance than separated extraction, which demonstrates class ties will improve relation extraction and the two methods are effective to learn class ties; (2) For Rank+AVE, surprisingly joint extraction does not keep up with separated extraction. For the second phenomenon, it may come from the strategy of \textbf{AVE} method to aggregate sentences. 
To make joint extraction, we will combine all the sentences containing the same entity tuple, however, not all sentences have the same relation, the fact is that one part of the sentences express one relation type and some will have another one. Simply averaging the sentence representations will hinder the model to learn the latent mapping from the sentences to the corresponding relation type, because averaging operation will gender redundant information from other unrelated sentences.  

\begin{figure}
\centering\includegraphics[width = 12cm]{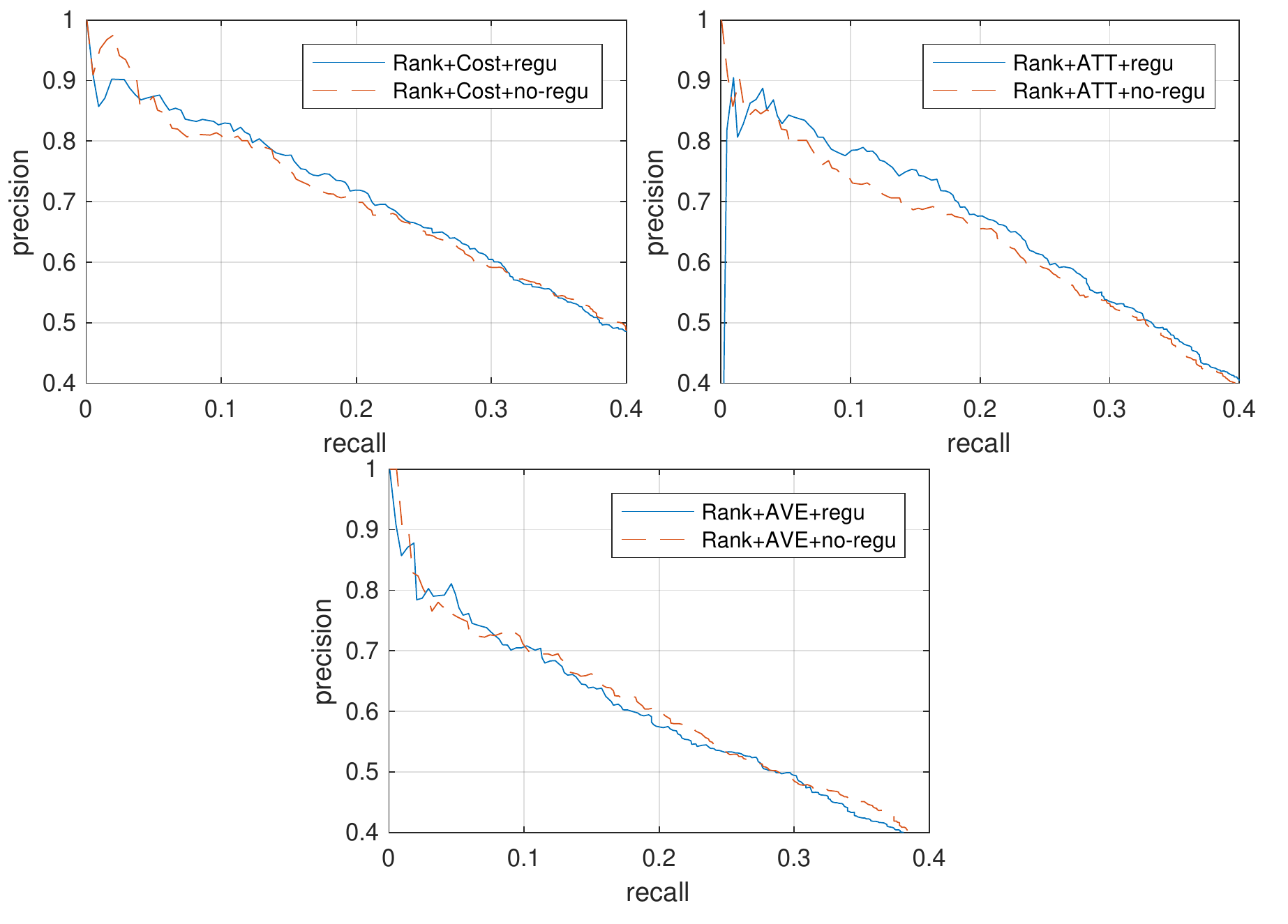}
\caption{Results for impact of regularization to model class ties.}
\label{fig_regu}
\end{figure}

\begin{table}
 \caption{Precisions for top $100$, $200$, $300$, $400$, $500$ and average of them for impact of regularization to model class ties}\label{tab_regu}
  \centering
  \begin{tabular}{p{2.7cm}|p{0.49cm}p{0.49cm}p{0.49cm}p{0.49cm}p{0.49cm}p{0.5cm}}
  \hline
  \textbf{P@N(\%)} & $\textbf{100}$ & $\textbf{200}$ & $\textbf{300}$ & $\textbf{400}$ & $\textbf{500}$ &\textbf{Ave.} \\ 
  \hline
  \textbf{R.+AVE+no-regu.}  & $78.0$  & $72.3$	&  $69.8$	& \textbf{66.5}	& \textbf{64.0} & $70.1$\\ 

  \textbf{R.+AVE+regu.} & \textbf{79.1} & \textbf{73.8}	& \textbf{70.4}	& $66.0$	& $63.1$ & \textbf{70.5}\\ 

  \hline
  \textbf{R.+ATT+no-regu.} & $84.6$ & $77.5$ & $72.9$ & $69.6$	& $68.0$ & $74.5$ \\ 
  \textbf{R.+ATT+regu.} & \textbf{86.8}	& \textbf{80.6}	& \textbf{78.4}	& \textbf{75.2}	& \textbf{71.1} & \textbf{78.4}\\ \hline

  \textbf{R.+Cost+no-regu.} & $85.7$ & $81.7$ & $80.1$ & $75.2$	& $71.3$ & $78.8$ \\ 
  \textbf{R.+Cost+regu.} & \textbf{86.8}	& \textbf{83.2}	& \textbf{81.1}	& \textbf{76.7}	& \textbf{73.5} & \textbf{80.3}	 \\ \hline
  \end{tabular}
\end{table}

$\bullet$ \textbf{Regularization.} To see the impact of regularization technique for modeling class ties, we compare the methods using regularization with the ones without using regularization. All variant models are in setting of joint extraction. The results are shown in Fig.~\ref{fig_regu} and Table~\ref{tab_regu}. From the results, we can see that after regularizing the learning of relations, the model performance can be further improved indicated by methods of Rank+Cost and Rank+ATT, which demonstrates the effectiveness of regularization to model class ties. We do not see many effects of regularization for method of Rank+AVE. Noises brought by averaging sentence embeddings may hinder the positive effects of regularization. 

\begin{figure}
\centering\includegraphics[width = 10cm]{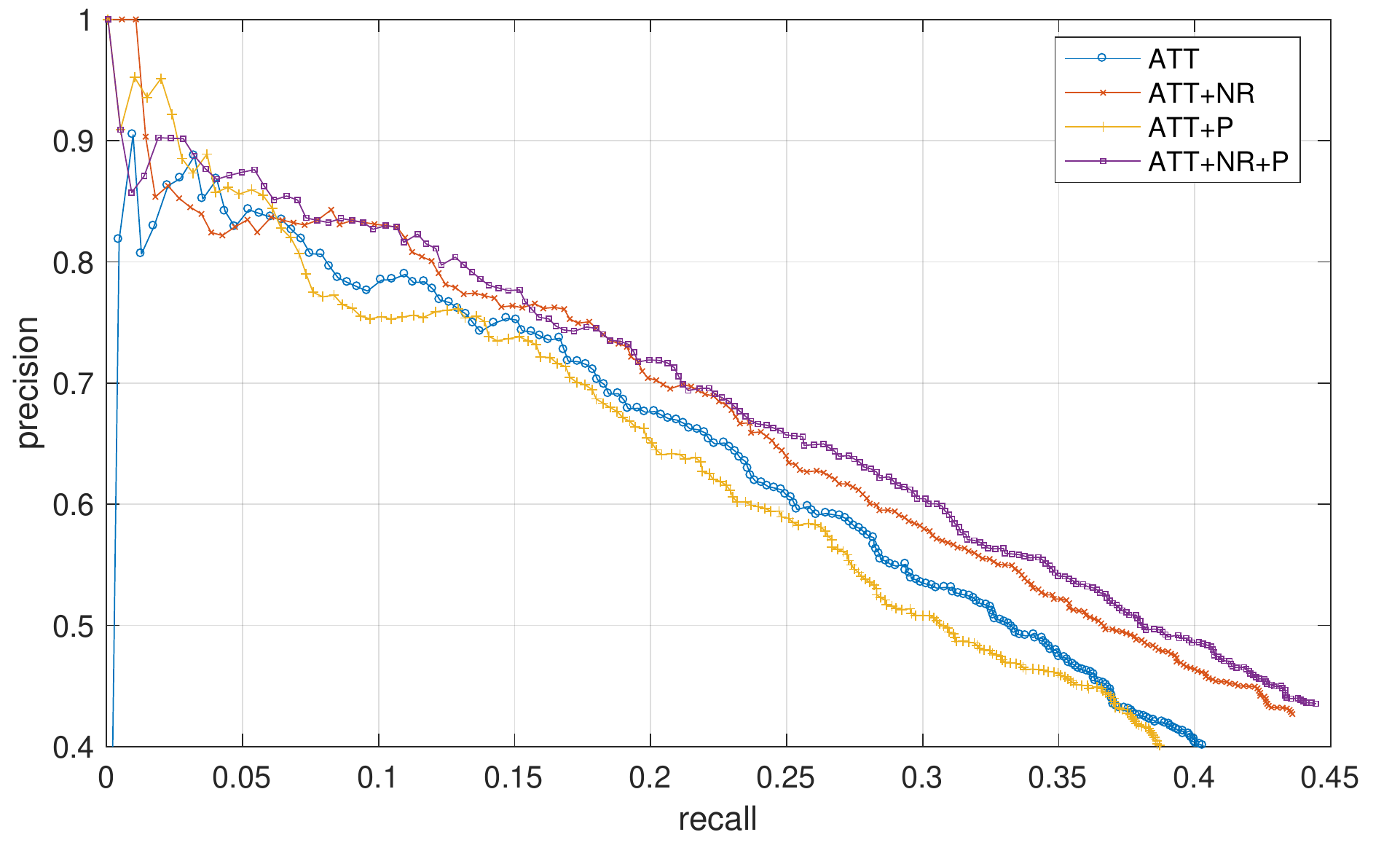}
\caption{Results for impact of cost-sensitive learning. ``ATT" means the loss function of Variant-2; ``ATT+NR" means only considering the cost of NR controlled by $\lambda$ ignoring the cost controlled by $\gamma$ based on Variant-2 and $\lambda$ is set to $0$; ``ATT+P" means considering the cost controlled by $\gamma$ based on Variant-2 ignoring the cost of NR and $\gamma$ is set to $1$; ``ATT+NR+P" is the loss function of Variant-3 and jointly considers the two kinds of costs mentioned above, $\lambda$ is set to $0$ and $\gamma$ is $1$.}
\label{class_imbalance}
\end{figure}

\begin{table}
 \caption{Precisions for top $100$, $200$, $300$, $400$, $500$ and average of them for impact of cost-sensitive learning.}\label{tab_cost_sen}
  \centering
  \begin{tabular}{p{1.8cm}|p{0.49cm}p{0.49cm}p{0.49cm}p{0.49cm}p{0.49cm}p{0.5cm}}
  \hline
  \textbf{P@N(\%)} & $\textbf{100}$ & $\textbf{200}$ & $\textbf{300}$ & $\textbf{400}$ & $\textbf{500}$ &\textbf{Ave.} \\ \hline
  \textbf{ATT}  & $86.8$  & $80.6$	&  $78.4$	& $75.2$	& $71.1$ & $78.4$\\ 
  \hline
  \textbf{ATT+NR} & $82.4$ & $84.3$	& $80.1$	& $76.2$	& $73.5$ & $79.3$\\ 
  \hline
  \textbf{ATT+P} & $85.7$ & $77.5$ & $75.6$ & $73.7$	& $69.9$ & $76.5$ \\ \hline
  \textbf{ATT+NR+P} & $86.8$  & $83.2$	& $81.1$	& $76.7$	& $73.5$ & $80.3$\\ \hline
  \end{tabular}
\end{table}
\subsection{Impact of Cost-sensitive Learning}\label{ImpactofCost-sensitiveLearning}
In this section, we conduct experiments to reveal the effectiveness of cost-sensitive learning to relieve the impact of NR for model training and model performance. For the loss function of $G_{[\mathrm{cost\_att]}}$, we have two parts for cost-sensitive learning: the first is the one penalized by  $\gamma$, and the second is the NR cost penalized by $\lambda$. Based on the loss function of Variant-3, we respectively relieve the cost controlled by $\gamma$ and the cost of NR controlled by $\lambda$ to see the impact of cost-sensitive learning. We will adopt P/R curves and precisions@N ($100$, $200$, $\cdots$, $500$) to show the model performances.

The results are shown in Fig.~\ref{class_imbalance} and Table~\ref{tab_cost_sen}. From the results, we can see that considering the cost controlled by $\gamma$ can sightly improve the performance in low recall range and considering the cost of NR controlled by $\lambda$ can boost the performance significantly. Considering both of the two kinds of costs can achieve the best performance. From these results, we can see that relieving NR impact is really important to improve the extraction performance.

\begin{figure}
\centering\includegraphics[width = 10cm]{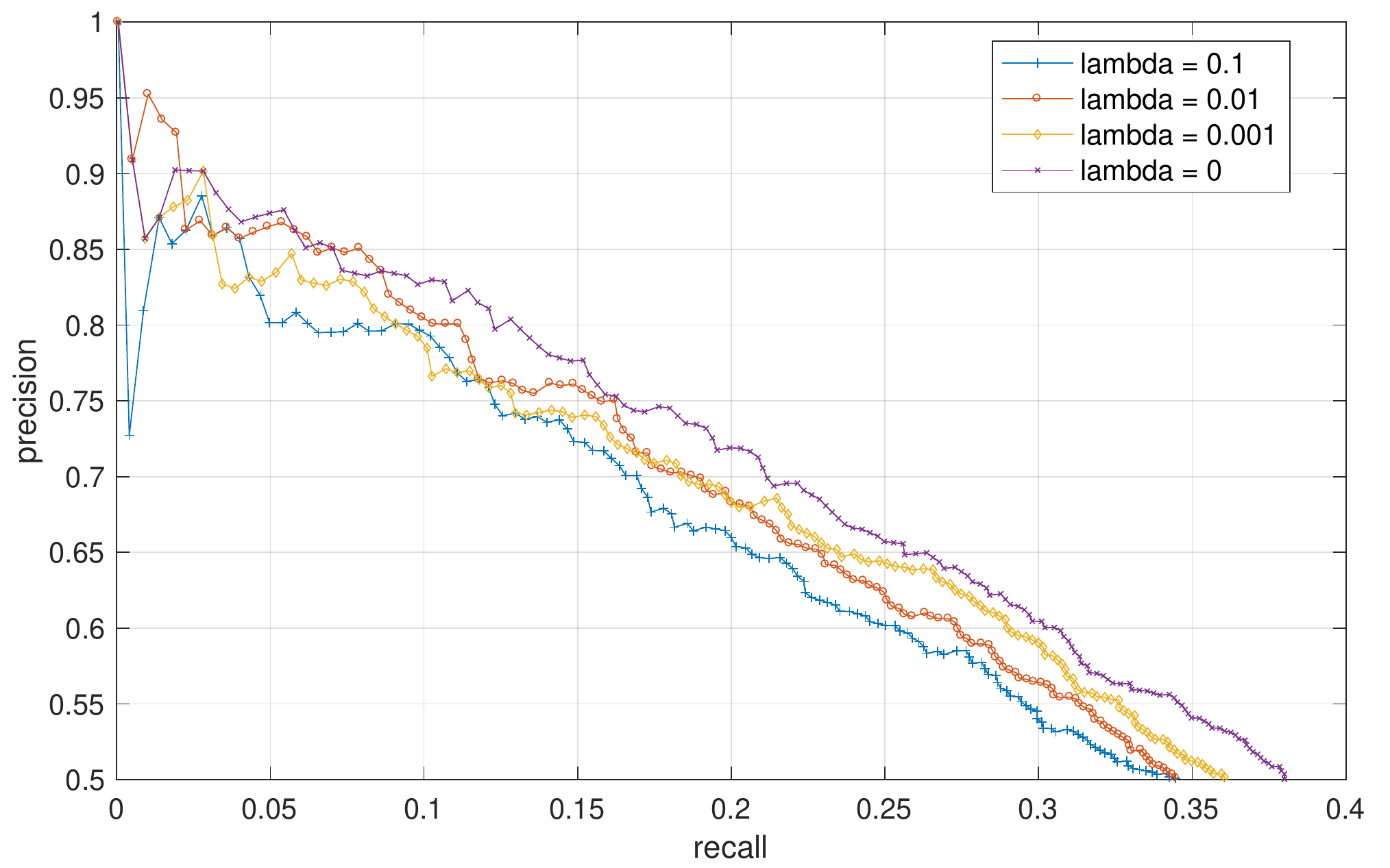}
\caption{Effect of $\lambda$ for model performance based on the loss function of Variant-3.}
\label{lambda}
\end{figure}

\subsection{Impact of NR}\label{ImpactOfNR}
From the discussion above, we can know that NR can have much significant impact for model performance, so in this section, we conduct more experiments to reveal the impact of NR cost controlled by $\lambda$ for model performance. 

$\bullet$ \textbf{Effect of $\lambda$ Penalty.}\ \ We conduct experiments on the choice of $\lambda$. Based on the loss function of Variant-3, we select $\lambda$ from $\{0, 0.001,0.01,0.1\}$ to see how much effect of NR can gender to the performance. We also adopt P/R curves and precisions@N ($100$, $200$, $\cdots$, $500$) to show the model performances. Models are set with joint extraction and regularization. The results are shown in Fig.~\ref{lambda} and Table~\ref{tab_lambda}. From the results we can find that when $\lambda$ becomes larger~(from $0$ to $0.1$), the model performance will decrease because NR will have more negative impact on model performance, so in order to achieve better model performance, the value of $\lambda$ should be set smaller. 

\begin{table}
 \caption{Precisions for top $100$, $200$, $300$, $400$, $500$ and average of them for impact of cost-sensitive learning.}\label{tab_lambda}
  \centering
  \begin{tabular}{p{1.6cm}|p{0.49cm}p{0.49cm}p{0.49cm}p{0.49cm}p{0.49cm}p{0.5cm}}
  \hline
  \textbf{P@N(\%)} & $\textbf{100}$ & $\textbf{200}$ & $\textbf{300}$ & $\textbf{400}$ & $\textbf{500}$ &\textbf{Ave.} \\ \hline

  \textbf{$\lambda$ = $0$}  & $86.8$  & $83.2$	&  $81.1$	& $76.7$	& $73.5$ & $80.3$\\ 

  \textbf{$\lambda$ = $0.001$} & $82.4$ & $82.2$	& $77.0$	& $73.9$	& $71.1$ & $77.3$\\ 

  \textbf{$\lambda$ = $0.01$} & $85.7$ & $84.3$ & $77.7$ & $75.7$	& $70.5$ & $78.8$ \\ 
  \textbf{$\lambda$ = $0.1$} & $85.7$  & $80.1$	& $76.3$	& $73.1$	& $68.6$ & $76.8$\\ \hline
  \end{tabular}
\end{table}

\begin{figure}
\centering\includegraphics[width = 10.cm]{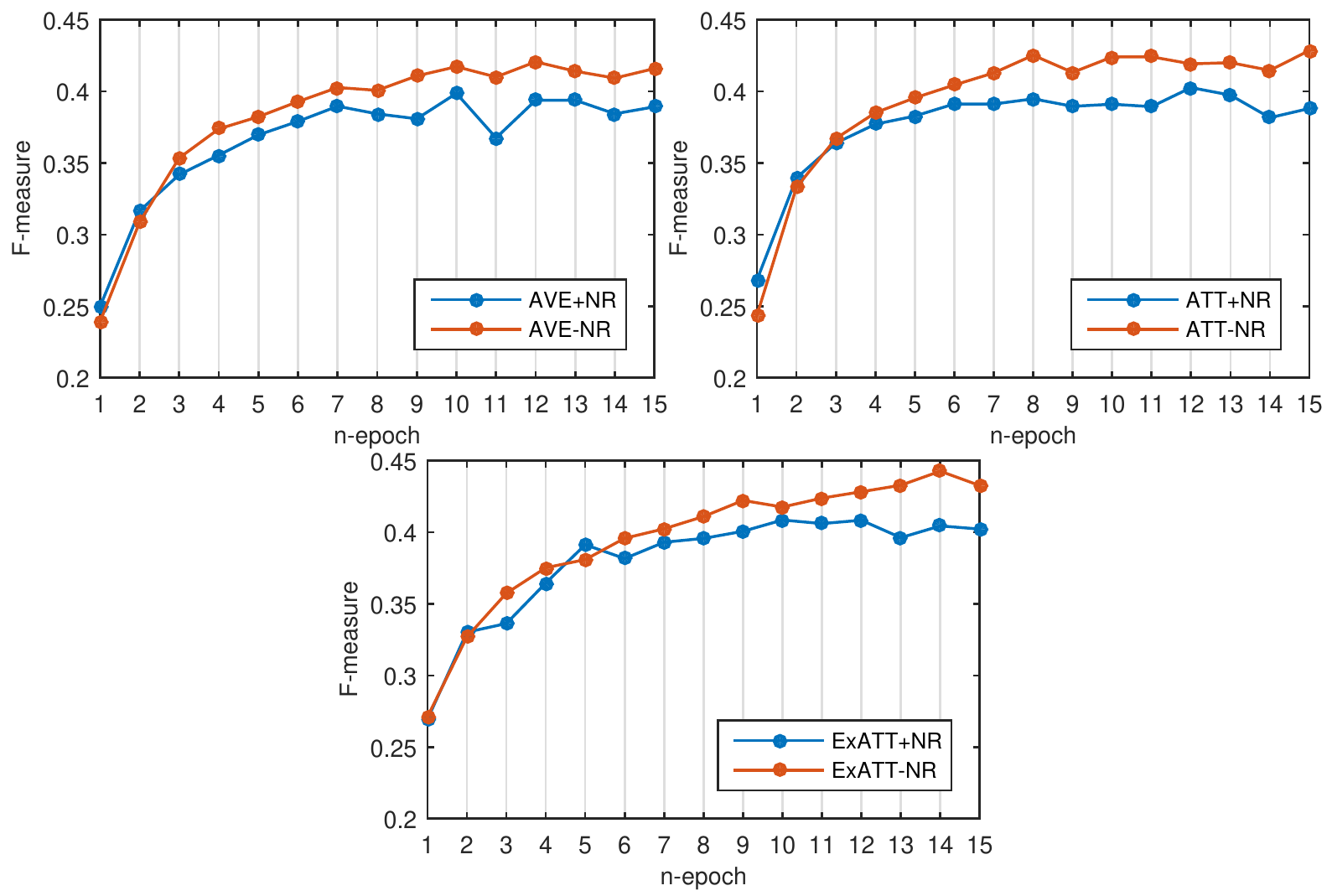}
\caption{\label{procedure} Impact of NR for model convergence. ``+NR" means not relieving NR impact with $\lambda$ of $1$; ``-NR" is opposite with $\lambda$ of $0$. ExATT is based on the loss function of Variant-3.}
\end{figure}
$\bullet$ \textbf{Effect of NR for Model Convergence.}\ \
Then we further evaluate the impact of NR for convergence behavior of our model in model training. Also with the three variant loss functions, in each iteration, we record the maximal value of F-measure
\footnote{$F = 2 * P * R / (P + R)$}
to represent the model performance at current epoch. Models are with setting of joint extraction but without regularization. Model parameters are tuned for $15$ times and the convergence curves are shown in Fig.~\ref{procedure}. From the result, we can find out: ``+NR" converges quicker than ``-NR" and arrives to the final score at the around $11$ or $12$ epoch. In general, ``-NR" converges more smoothly and will achieve better performance than ``+NR" in the end.


\section{Conclusion and Future Works}
In this work, we propose a ranking based cost-sensitive multi-label learning for distant relation extraction aiming to leverage class ties to enhance relation extraction and relieving class imbalance problem. To exploit class ties between relations to improve relation extraction, we propose a general ranking based multi-label learning framework combined with convolutional neural networks, in which ranking based loss functions with regularization technique are introduced to learn the latent connections between relations. Furthermore, to deal with the problem of \emph{class imbalance} in distant supervision relation extraction, we further adopt cost-sensitive learning to rescale the costs from the positive and negative labels. In the experimental study, we further do experiments to analyze the effectiveness of our novel cost-sensitive ranking loss functions. The evaluation experiments on the effectiveness of regularization have further be conducted.

In the future, we will focus on the following aspects: (1) Our method in this paper considers pairwise intersections between labels, so to better exploit class ties, we will extend our method to exploit all other labels' influences on each relation for relation extraction, transferring \emph{second-order} to \emph{high-order}~\cite{zhang2014review}; (2) We will regard the task of distant supervision relation extraction as a multi-instance based learning-to-rank problem, and will take the view from learning-to-rank to design the algorithms and combine other advanced tricks from information retrieval field; (3) What effects will entity pairs take to the relation extraction performance? Can we use a general entity pair replacement ($e_1$, $e_2$) to represent all entity pairs? Answering the two problems may help the transfer learning of RE systems.

\section*{Acknowledgment}
This work was supported by the National High-tech Research and Development Program (863 Program)
(No. 2014AA015105) and National Natural Science Foundation of China (No. 61602490).

\section*{References}

\bibliography{elsarticle-template}
\bibliographystyle{elsarticle-num}

\end{document}